\documentclass[]{article}

\usepackage{PRIMEarxiv}

\usepackage[utf8]{inputenc} % allow utf-8 input
\usepackage[T1]{fontenc}    % use 8-bit T1 fonts
\usepackage{hyperref}       % hyperlinks
\usepackage{url}            % simple URL typesetting
\usepackage{booktabs}       % professional-quality tables
\usepackage{amsfonts}       % blackboard math symbols
\usepackage{nicefrac}       % compact symbols for 1/2, etc.
\usepackage{microtype}      % microtypography
\usepackage{lipsum}
\usepackage{fancyhdr}       % header
\usepackage{graphicx}       % graphics
\graphicspath{{media/}}     % organize your images and other figures under media/ folder
\usepackage{amsthm}
\usepackage{amssymb}
\usepackage{amsmath}
\usepackage[utf8]{inputenc}
\usepackage{lmodern}
\usepackage{xcolor}
\usepackage{graphicx}
\usepackage{subcaption}
\usepackage{booktabs}
\usepackage{natbib}

\title{Bayesian Entropy Neural Networks for Physics-Aware Prediction
}

\author{
  Rahul Rathnakumar, Yongming Liu \\
  School for Engineering of Matter, Transport and Energy \\
  Arizona State University \\
  Tempe \\
  \texttt{\{rrathnak, yongming.liu\}@asu.edu} \\
   \And
  Jiayu Huang, Hao Yan \\
  School of Computing and Augmented Intelligence \\
  Arizona State University \\
  Tempe \\
  \texttt{\{jhuan178, haoyan\}@asu.edu} \\
}

\begin{document}
\maketitle

\begin{abstract}
This paper addresses the need for deep learning models to integrate well-defined constraints into their outputs, driven by their application in surrogate models, learning with limited data and partial information, and scenarios requiring flexible model behavior to incorporate non-data sample information. We introduce Bayesian Entropy Neural Networks (BENN), a framework grounded in Maximum Entropy (MaxEnt) principles, designed to impose constraints on Bayesian Neural Network (BNN) predictions. BENN is capable of constraining not only the predicted values but also their derivatives and variances, ensuring a more robust and reliable model output. To achieve simultaneous uncertainty quantification and constraint satisfaction, we employ the method of multipliers approach. This allows for the concurrent estimation of neural network parameters and the Lagrangian multipliers associated with the constraints. Our experiments, spanning diverse applications such as beam deflection modeling and microstructure generation, demonstrate the effectiveness of BENN. The results highlight significant improvements over traditional BNNs and showcase competitive performance relative to contemporary constrained deep learning methods.
\end{abstract}

% keywords can be removed

%% main text
\section{Introduction}
\label{sec:introduction_BENN}
Neural networks have achieved remarkable success in various domains, ranging from image recognition to natural language processing \citep{goldberg2016primer, khan2020survey}. Despite these advancements, traditional deep learning approaches often fall short in scenarios that demand adherence to specific constraints \citep{zhu2019physics, hosseini2015deep} or a deeper understanding of model uncertainty \citep{wilson2020bayesian}. This limitation becomes particularly pronounced in fields such as surrogate modeling, learning with limited or partial data, and applications where flexible adaptation to non-sample-based information is crucial. For surrogate models, the neural network model is expected to conform to behavior governed by the system model, and without enough data across the domain of interest, neural network performance falls short. Integrating domain knowledge into these neural networks remains challenging. Existing works on integrating domain knowledge use either Bayesian Networks, Physics-Informed Neural Networks (PINNs) or more general regularization schemes. 
\par
Bayesian Networks are probabilistic graphical models that represent a set of variables and their conditional dependencies via a directed acyclic graph (DAG). They are particularly useful for modeling complex systems where various factors influence each other in a probabilistic manner. In the context of physics and data-driven models, Bayesian Networks can be employed to incorporate prior knowledge and physical laws as constraints, ensuring that the models adhere to known scientific principles. \citet{Wang_2020_BayesEnt} presents a method for information fusion combining the maximum entropy (ME) method with the classical Bayesian network, termed the Bayesian-Entropy Network (BEN). This method is particularly adept at handling various types of information for classification and updating, such as point data, statistical information, and range data. The BEN method extends the Bayesian approach by integrating additional information in the form of constraints into the entropy part, while the Bayesian part handles classical point observation data. This integration allows for a more comprehensive approach to modeling and decision-making. The BEN method demonstrates a flexible adaptation to non-sample-based information, making it particularly relevant in fields like surrogate modeling and learning with limited or partial data. For example, \citet{Wang_2021_BEGP} applies this method to the Gaussian Process model by adding information as constraints. Another work that uses such surrogate modeling techniques in conjuction with Bayesian Networks is by \citet{LU2022108290} for the risk assessment of a power distribution system. In this work, the authors use the physical properties of the individual system components and the network topology to derive a physics-based fragility analysis for the whole system. Using Bayesian Networks, they calculate the time-dependent failure rate function of the system. In this paper, such modeling efforts are extended to apply to neural network models, showcasing its potential in tasks such as information constraints for high-dimensional problems such as microstructure generation, but also in controlling the predicted uncertainty, which is an important avenue of research. The reason for the latter is motivated by wanting to output high uncertainty for predictions outside the training domain, or if the information constraints are conflicting, as will be shown in the later stages of this paper.
\par
It is not widely known in the community that artificial neural networks have been effectively applied to physical systems for problem-solving. For instance, \citet{lagaris1998artificial} utilized a multi-layer perceptron (MLP) to address partial differential equations (PDEs). This approach involves using the MLP to generate trial solutions that meet initial and boundary conditions by minimizing an energy function derived from the differential equation. However, one of the limitations of this approach to incorporating physics information is that it is sensitive to mesh density. This also implies that there are limitations on the dimensions of the problems that can be solved. Recently, \citep{Sirignano2017DGMAD} proposed a mesh-free, deep learning based technique that takes inspiration from Galerkin method for solving PDEs that is accurate at high dimensions, and alleviates this issue. The idea here is to use a neural network to approximate the solution instead of basis function combinations. While these methods are focused on solving differential equations, it is still relevant to scientific machine learning and building surrogate models for physical systems. A landmark contribution in this line of work was  \citet{Raissi_2019}, which presented physics-informed neural networks (PINNs) that directly incorporate physical laws in the form of differential equations, directly into the loss function of the neural network. This is achieved by constructing a composite loss function that not only measures the difference between predictions and data but also quantifies the deviation from the specified physical laws. PINNs are specialized models that have been shown to model differential equation constraints accurately using regularization. However, these types of techniques focus on compelling PDE and ODE constraints, which limits their use for general knowledge constraints, which is the major focus of this paper. 
\par
In the third area, Posterior-regularised BNN \citep{Jiayu_2022} and prior-regularized BNN \citep{Yang_2020_Interpretable, sam2024bayesian, tran2022need} are contemporary papers that are related to this work. \citet{Jiayu_2022} focuses on embedding soft and hard knowledge constraints into the posterior, offering enhanced model robustness and adaptability. \citet{Yang_2020} modifies the prior, where the prior is optimized to comply with the constraints. Where OC-BNNs \citep{Yang_2020_Interpretable} primarily focus on defining equality and inequality constraints by multiplying likelihoods and priors in the probability space, and PR-BNNs directly regularize the posterior predictive distribution to comply with the constraint. Modifying the prior does have it's advantages- For instance, by using a domain knowledge loss on unlabeled samples, \citet{sam2024bayesian} modifies a low-rank gaussian prior into an informative prior. The idea is that the learned informative prior obtained during the pre-training stage should transfer its properties to the posterior, without having to add a loss to the posterior, which needs sampling. But it isn't clear that the prior can be specified correctly and easily for complex problems - the predictive distribution is more intuitive to work with when the modeler wants to incorporate constraints. Moreover, the fact that we already require posterior samples for evaluating the KL-Divergence means that the addition of a constraint term on the posterior is just another term that needs to be evaluated along side the data loss. However, \citet{sam2024bayesian} argues that since the informative prior is low-rank Gaussian, the computational overhead of using an informative prior isn't higher than using an isotropic Gaussian. Be it using posterior regularization or prior regularization, the common theme across these papers is the use of loss functions to enforce compliance constraints on model behavior. However, these papers have also not studied extensively the variety of constraints that are explored in this paper. For example, while PR-BNN can decide the amount of constraints penalty be added in the model, it lacks the power to balance multiple knowledge constraints, especially if they are conflicting. Furthermore, the work presented in this paper also studies variance constraints and constraints applied to high dimensional image generation problems, where the constraints are not obvious and highlights some nuances in constraint modeling for neural network-based tasks in the real world. 
\par
In this paper, we propose a framework that generalizes existing approaches by integrating a broad range of constraints and automatically determining their strength using the Maximum Entropy (MaxEnt) principle. The core of Bayesian Entropy Neural Networks (BENN) is their ability to combine the MaxEnt principle with the Bayesian Neural Network framework. This results in a model that offers constrained and uncertain predictions while remaining computationally feasible. Bayesian Neural Networks (BNNs) are founded on the principles of Bayesian inference and provide an elegant solution to the over-reliance on point estimates inherent in traditional neural networks. By leveraging posterior distributions over parameters, BNNs offer a more nuanced view of model uncertainty, essential for many real-world applications. However, the practical implementation of Bayesian updating in neural networks is challenged by computational limitations, especially with large parameter spaces. To address this issue, we use Variational Inference (VI), an approach that significantly reduces computational overhead by approximating posterior distributions. This approximation is crucial for applying BNNs in more complex and realistic scenarios. Works such as \citet{Wang_2021} use a sequential optimization approach to iteratively find model parameters for the Gaussian Process and the Lagrange multipliers. Neural networks, however, provide backpropagation machinery to handle optimization under constraints. Therefore, this work utilizes the Modified Differential Method of Multipliers (MDMM) to efficiently learn constraint weights.
\par
The rest of the paper is organized as follows:
\begin{enumerate}
    \item Section \ref{sec:method_BENN} provides background information on Bayesian Neural Networks, the Maximum Entropy method, and concludes with a description of the Bayesian Entropy Neural Network (BENN) framework.
    \item Section \ref{sec:experiments} covers three experiments: 
    \begin{enumerate}
        \item 1D Regression: Implementing value, derivative, and predictive variance constraints.
        \item Beam Deflection: Demonstrating the performance of the BENN approach in a beam deflection problem, where part of the domain is observed.
        
        \item Microstructure Generation: Showcasing the benefit of constrained learning to enhance the performance of microstructure generation using suitable constraints on the microstructure properties.
    \end{enumerate}
    \item Section \ref{sec:Conclusion_BENN} summarizes the work done in this paper and highlights the importance of the main results obtained.
\end{enumerate}
\section{Method}
\label{sec:method_BENN}
Here, we will present the major methodology of the proposed BENN method. We will start with a brief overview of Bayesian Neural Networks (BNNs) in Section \ref{subsec:BNN}. We will then discuss the integration of maximum entropy principles into Bayesian inference and introduce the Bayesian Entropy Neural Network (BENN) framework in Section \ref{subsec:BayesEnt}.

\subsection{Overview of Bayesian Neural Networks}
\label{subsec:BNN}
Bayesian Neural Networks arose out of a need to have a principled alternative to point estimates of parameters in standard neural networks. Given a neural network \(f(\theta)\), and dataset \( D = {(x_i, y_i)}_{i=1}^n\), one can formulate the process as updating a prior belief \(p(\theta)\) using a likelihood \(p(D|\theta)\), to obtain a posterior \(p(\theta|D)\):
\begin{equation}
    p(\theta|D) = \frac{p(D|\theta)\cdot p(\theta)}{p(D)}
\end{equation}
In the above equation, the denominator \(p(D)\) is the marginal likelihood of the data, that is obtained after marginalization of the parameter \(\theta\). While the Bayes rule is valid for any family of models, implementing updating schemes for neural networks presents a challenge. The dimension of the parameter set \(\theta\) can be large, leading to computational difficulties in calculating the marginal likelihood. Moreover, the goal of inferring parameters is to predict on test samples \(x^*\). However, one again confronts a need to marginalize the parameters using \( p(y^*|x^*,D) = \int p(y^*|x^*, \theta) p(\theta|D) d\theta \) to get an averaged prediction. Sampling approaches such as Markov Chain Monte Carlo (MCMC) are often used in smaller models to get accurate estimates of the true posterior distributions. However, for neural networks, this sampling approach is disfavored due to excessive computational requirements. Instead, variational inference techniques are used.
\par
VI approximates the posterior by finding a proposal distribution \(q_\phi(\theta)\) that minimizes the KL-divergence between the two. The KL divergence equation is given by:
\begin{equation}
    \text{KL}(q_\phi(\theta)||p(\theta|D)) = E_q[\log(q_\phi(\theta))] - E_q[\log(p(\theta|D))]
\end{equation}
From the Bayes rule, it is seen that the second term is still dependent on the marginal likelihood, which is intractable. However, \(p(D)\) is independent of the proposal distribution \(q_\phi(\theta)\). Taking advantage of this fact, variational methods maximize an equation that is proportional to the KL-divergence, the Evidence Lower Bound (ELBO):
\begin{equation}
\begin{split}
\text{ELBO}(q) & = E_q[\log(p(D|\theta))] + E_q[\log(p(\theta))] - E[\log(q_\phi(\theta))] \\
        & = E_q[\log(p(D|\theta))] - \text{KL}(q_\phi(\theta)||p(\theta))
\end{split}
\end{equation}
The two components present in the final form consist of an expected log-likelihood term that is data-dependent, and a term that encourages the proposal distribution to not deviate from the prior. 
\par
The ELBO equation has gradients that can be estimated using Monte Carlo sampling of the parameter \(\theta^i\) from the approximated posterior after each iteration of the training loop.
\begin{equation}
    L_{VI} = \sum_{i=1}^{n} log(q_\phi(\theta^i)) - P(\theta^i) - P(D|\theta^i)
\end{equation}
Stochastic gradients are accomplished using a reparameterization:
\begin{equation}
    z = \mu + \epsilon \cdot \exp\left(\frac{1}{2} \log(\sigma^2)\right)
\end{equation}
\par
The variational approximation can also be used to generate samples from a latent encoding \(q(z)\) of an input \(x\), using the commonly used Variational Auto-Encoder model (VAE). Further details on this experimental setup are given in Section~\ref{subsec: Microstructure_VAE}.
\subsection{Incorporating Constraint Information in Bayesian Entropy Method}
\label{subsec:BayesEnt}
The method of maximum entropy, seen in \citep{Giffin_2007} was proposed by Jaynes to perform inference of model parameters subject to constraints. In other words, MaxEnt picks parameters that maximize the entropy of its distribution insofar as the constraints allow. As a corollary, it is also obtained that given no information about the parameters, a flat distribution is the one that maximizes entropy. This line of reasoning suggests that this method fits well within the Bayesian updating paradigm, as constraints are a form of information about the parameters.
\par
If we have a joint prior \(p((\theta), x)\) on a model given some information \(F\) about the model parameters, we can maximize the entropy between the prior and the posterior update considering the constraints as an optimization problem:
\begin{equation}
    \begin{split}
        \text{argmin}_{\theta} -\int P(\theta, x_c, x) log \frac{P(\theta, x_c, x)}{P(\theta, x)} dx d\theta \\
        \text{s.t.} \int dx d\theta p(\theta, x_c, x) &= 1 \\
        \int p(x_c,x,\theta)f(\theta) d\theta dx &= F
    \end{split}
\end{equation}
The above equation formalizes the entropy maximization problem subject to a constraint on the value of p and a normalization constraint on p. The solution is derived using the method of Lagrangian multipliers. If one extends the above-constrained system to m constraints, the objective is to solve for \(m + 1\) multipliers:
\begin{equation}
    \begin{split}
        L(p, \lambda, \nu) &= \int p_i log(p_i) dp - \sum_{k=1}^m(\lambda_k\int(p_i f_k(x_i) - F_k) dp - \nu\int(p_i - 1) dp
    \end{split}
\end{equation}
The solution takes the form of an exponential function normalized by a partition function Z:
\begin{equation}
    \begin{split}
        p(\theta|x, x_c) &= \frac{1}{Z(\lambda_1, ... , \lambda_m)}\cdot \exp (-\sum_{k=1}^{m} \lambda_k f_k(x)) \\
        Z(\lambda_1, ... , \lambda_m) &= \int \exp(-\sum_{k=1}^m \lambda_k f_k(x))
    \end{split}
\label{eq: eqn8}
\end{equation}
From the above description, it is seen that the inference procedure requires a prior that captures no new information apart from the constraints and the data. Therefore, one choice to model this is to use a uniform prior. For a distribution \( q_{\phi}(\theta) \) and a uniform prior distribution \( p(\theta) \), the Kullback-Leibler (KL) divergence from \( q_{\phi}(\theta) \) to \( p(\theta) \) can be calculated as:
% \end{proposition}
\begin{equation}
D_{KL}(q_{\phi}(\theta) || p(\theta)) = \int q_{\phi}(\theta) \log \frac{q_{\phi}(\theta)}{p(\theta)} \, dw
\end{equation}
With \( p(\theta) = \frac{1}{|\Theta|} \) for all \( \theta \in \theta \), the KL divergence simplifies to:
\begin{equation}
D_{KL}(q_{\phi}(\theta) || p(\theta)) = \int q_{\phi}(\theta) \log (q_{\phi}(\theta) \cdot |\Theta|) \, d\theta
\end{equation}
This can be decomposed into:
\begin{equation}
D_{KL}(q_{\phi}(\theta) || p(\theta)) = \int q_{\phi}(\theta) \log q_{\phi}(\theta) \, d\theta + \log |\Theta| \int q_{\phi}(\theta) \, d\theta
\end{equation}
Given that \( q_{\phi}(\theta) \) is a probability distribution, the second integral evaluates to 1, yielding:
\begin{equation}
D_{KL}(q_{\phi}(\theta) || p(\theta)) = \int q_{\phi}(\theta) \log q_{\phi}(\theta) \, d\theta + \log |\Theta|
\end{equation}
The integral represents the negative of the entropy of \( q_{\phi} \), denoted as \( -H(q_{\phi}) \). Therefore, the KL divergence can be rewritten as:
\begin{equation}
D_{KL}(q_{\phi}(\theta) || p(\theta)) = -H(q_{\phi}) + \log |\Theta|
\end{equation}
Since \( \log |\Theta| \) is constant, the negative KL divergence is effectively proportional to the entropy of the variational distribution. This has indeed been concurrently investigated by \citet{de2023maximum}, where the model uses just the entropy of the proposal distribution for KL-divergence minimization. This is motivated by the above reasoning using the MaxEnt principle. Effectively, MaxEnt ensures that the prior assumptions are minimal, as the uniform prior contributes only a constant term to the KL divergence, making no further assumptions about the distribution of the parameters. 

\subsection{Bayesian Entropy Neural Networks}
\label{subsec:BENN}
Building on the MaxEnt principle and Bayesian Neural Networks (BNNs), this section introduces Bayesian Entropy Neural Networks (BENN). The previous section modeled constraints on the posterior distribution. Now,   constraint enforcement on both the posterior predictive and the parameters is demonstrated using this framework. 
\par
Given a Bayesian Neural Network \(f_{BNN}(\theta)\), prediction \(\hat{y}\) and a constraint function \(g\), the learning objective can be formulated a:
\begin{equation}
\begin{aligned}
\theta^* = & \, \text{argmin}_\theta \left( KL(q_\phi(\theta) \| p(\theta)) - E_{q_\phi(\theta)}[\log p(D \mid \theta)] \right) \\
 & \text{s.t.} \quad g(x) \le 0 \\ 
                    & h(x) = 0 \\
\end{aligned}
\end{equation}
For brevity, we write \( L_{VI} = KL(q_\phi(\theta)||p(\theta)) - E_q[\log p(D|\theta)]\).
Using Lagrange multipliers, the loss function can be formulated as:
\begin{equation}
    L = \text{argmin}_\theta L_{VI} + \lambda \cdot h(x) + \mu \cdot g(x)
\end{equation}
The above hard-constrained problem must strictly satisfy the inequality and equality constraints while also minimizing the loss function. In the case of a soft constraint, there can be violations allowed on the constraints using a slack variable \(\xi \) on the constraint infeasibility term:
\begin{equation}
    L = \text{argmin}_\theta L_{VI} + \lambda \cdot h(x) + \mu \cdot (g(x) - \xi^2)
\end{equation}
It is easy to see that this regularization form can theoretically satisfy both the constraint and the data, provided the neural network has enough parameters. However, a loss that includes the constraint as a regularizer, as written above, may not always result in a local minimum but can also be a stationary point. Oftentimes, practitioners perform a grid search on the values of the constraint weights to select one that maximizes a metric of interest or use a value to assign importance to the constraint compared to the data. To improve this, a more principled, robust updating scheme is utilized for parameters and the Lagrangian multipliers.
The Modified Differential Method of Multipliers approach (MDMM), proposed by \citet{Platt_1987}, is found to be well-suited, both from a theoretical perspective, as detailed in the original paper that proposed this optimization technique, and from a practical standpoint, for fitting the constraint weights. MDMM adds an extra damping term, similar to penalty methods, to the loss function and has the following form:\begin{equation}
\begin{split}
        \text{argmin}_\theta L(p, \lambda, c) &= \text{argmin}_\theta L_{VI} + \lambda h(x) + \frac{c_1}{2} h(x)^2 + \mu (g(x) - \xi^2) + \frac{c_2}{2} g(x)^2
\end{split}
\end{equation}
From Section~\ref{subsec:BayesEnt}, it is seen that the constrained optimization problem leads to an updated posterior of the form given by Equation~\ref{eq: eqn8}. Note that BENN gives us an easy way to sample from the posterior distribution, so one can extract \(N\) samples from the posterior to evaluate the constraint. 
\par
Compared to \citep{Wang_2021} work on Bayesian Entropy Gaussian Process (BEGP), the BENN leverages the high dimensional representation capacity of neural networks, and uses the same underlying principle to enforce constraints. Moreover, the optimization framework obtained as a result of the backpropagation algorithm makes it possible to simultaneously optimize for both the model parameters and the Lagrangian multipliers.  Gaussian Processes, however, have inherent advantages in uncertainty quantification that Bayesian Neural Networks do not, and this paper attempts to incorporate the principles from MaxEnt to demonstrate constraining predictive variance as well. 
\section{Experiments}
\label{sec:experiments}
\subsection{1D-Regression}
In this section, the data, architecture, and loss used for the regression problem are first described. Following this, a demonstration of value and derivative constraints is shown. Finally, demonstrations to constrain the behavior of the predictive variance for 1-D regression are shown.
\par
The dataset for the 1D-regression problem models a polynomial function with added noise and is defined over two distinct regions. In both regions, the function is given by:
\begin{equation}
y = p_0 x^2 + p_1 x + p_2 + 0.15 \sin(2 \pi x) + \mathcal{N}(0, \sigma)
\end{equation}
where the coefficients are \(p_0 = 0.2\), \(p_1 = 0.05\), and \(p_2 = 0.01\). The first region, from \(x = - 3\) to \(x = - 2\), has a noise level with a standard deviation of \(\sigma =0.01\). The second region, from \(x = 2.0\) to \(x = 3.0\), has a higher noise level with standard deviation \(\sigma = 0.1\). The training data \(X_{train}\) and corresponding labels \(Y_{train}\) consist of samples from both regions.
\par
The architecture of the neural network used in these experiments is a Bayesian Neural Network (BNN). The network comprises one hidden layer with 100 units and employs the Rectified Linear Unit (ReLU) as the activation function. The output layer of the network is designed to produce two outputs: the mean prediction of the target variable and a parameter termed as \(\log(\text{noise})\) or \(\sigma\), which represents the log-transformed noise level of the predictions. For the expected likelihood loss, it is assumed that the data generated is from a Gaussian, and a negative log-likelihood loss is minimized:
\begin{equation}
\mathcal{L}(\mathbf{y}, \hat{\mathbf{y}}, \sigma) = \frac{1}{N} \sum_{i=1}^{N} \left( \frac{(y_i - \hat{y}_i)^2}{2\exp(\sigma)} + \frac{\sigma}{2} \right)
\end{equation}
Here, \(\mathbf{y}\) represents the actual values, \({\mathbf{\hat{y}}}\) denotes the predicted mean values, \(N\) is the number of data points, and \(\sigma\) is the log-noise parameter. This loss function not only penalizes the deviation of predictions from the actual values but also incorporates the uncertainty associated with the predictions as indicated by \(\sigma\). 
\par
From the point of view of Maximum Entropy (MaxEnt), one seeks to make minimal assumptions about the nature of the model before training. In the Bayesian framework, this can be achieved by using a prior that is uniform over the space of parameters.
\subsubsection{Incorporating Value and Derivative Constraints to the Prediction}
\label{subsubsec:1dreg_valDerivConstraints}
\begin{figure}[htbp]
    \centering
    \begin{subfigure}[b]{0.49\textwidth}
        \includegraphics[width=\textwidth]{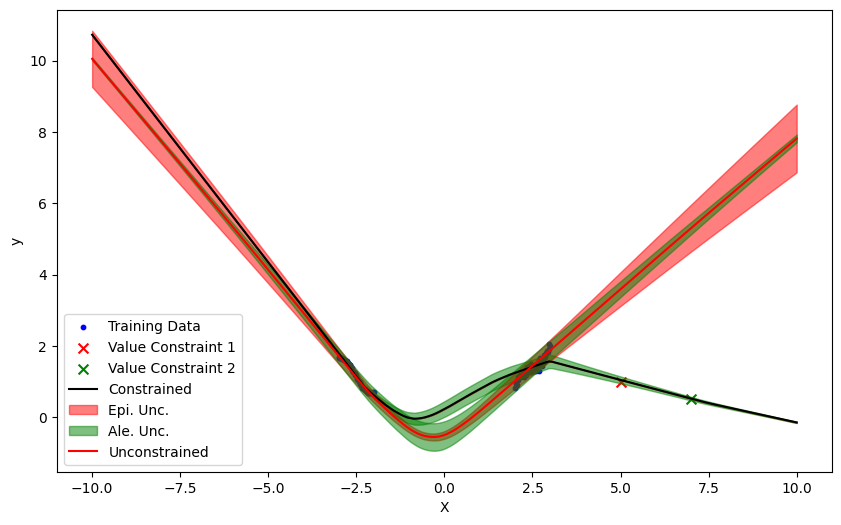}
        \caption{Value constraints set with \(\sigma_{ale} = 0.0\)}
        \label{fig:1d_regressionPlots_a}
    \end{subfigure}
    \hfill
    \begin{subfigure}[b]{0.49\textwidth}
        \includegraphics[width=\textwidth]{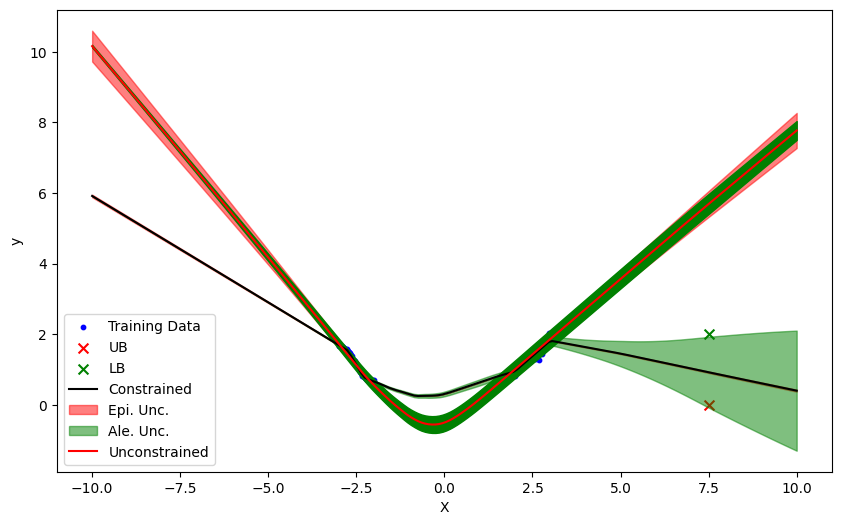}
        \caption{Conflicting value constraints with \(\sigma_{ale} = 2.0\)}
        \label{fig:1d_regressionPlots_b}
    \end{subfigure}
    \hfill
    \begin{subfigure}[b]{0.49\textwidth}
        \includegraphics[width=\textwidth]{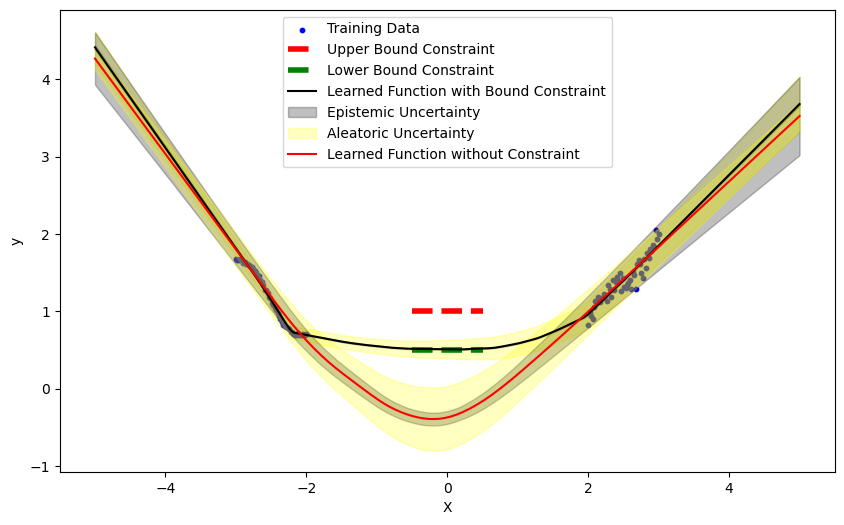}
        \caption{Bound constraint in x = [-0.5, 0.5]}
        \label{fig:1d_regressionPlots_c}
    \end{subfigure}
    \hfill
    \begin{subfigure}[b]{0.49\textwidth}
        \includegraphics[width=\textwidth]{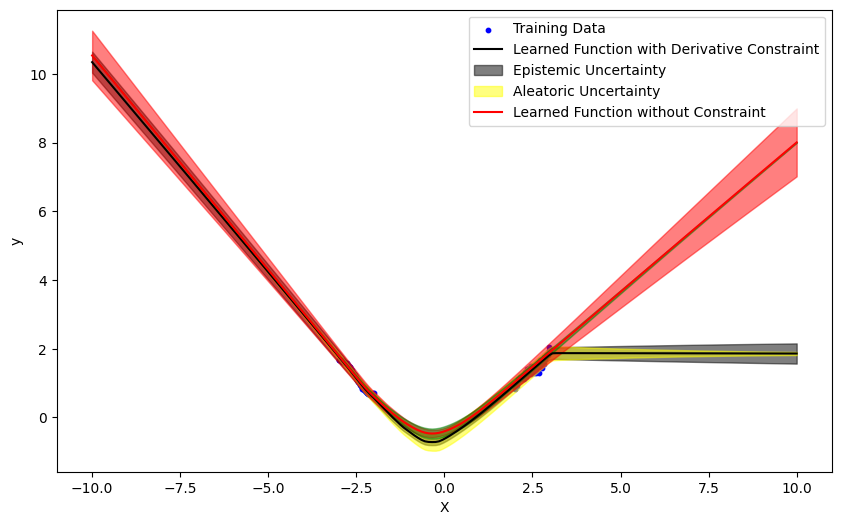}
        \caption{Derivative constraint \(\frac{dy}{dx} = 0, x = {9.5}\)}
        \label{fig:1d_regressionPlots_d}
    \end{subfigure}
    \hfill
    \caption{1-D constrained regression demonstrations with value, derivative and bound constraints}
    \label{fig:1d_regressionPlots}
\end{figure}

\begin{figure}[htbp]
    \centering
    % First col
    \begin{subfigure}[b]{0.32\textwidth}
        \includegraphics[width=\textwidth]{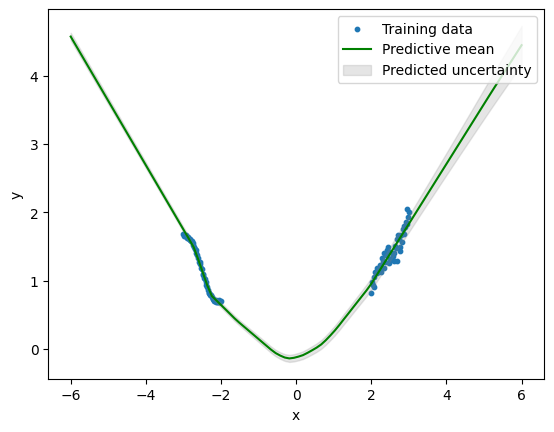}
        \label{fig:arbitraryVariance_low}
        \caption{Low variance outside the training data}
    \end{subfigure}
    \hfill
    \begin{subfigure}[b]{0.32\textwidth}
        \includegraphics[width=\textwidth]{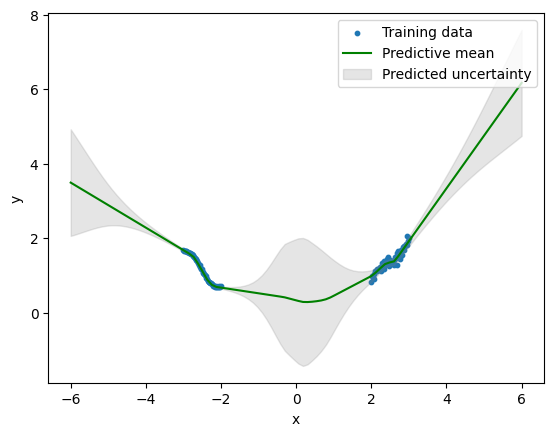}
        \label{fig:arbitraryVariance_med}
        \caption{Moderate variance outside the training data}
    \end{subfigure}
    \hfill
    % Second row
    \begin{subfigure}[b]{0.32\textwidth}
        \includegraphics[width=\textwidth]{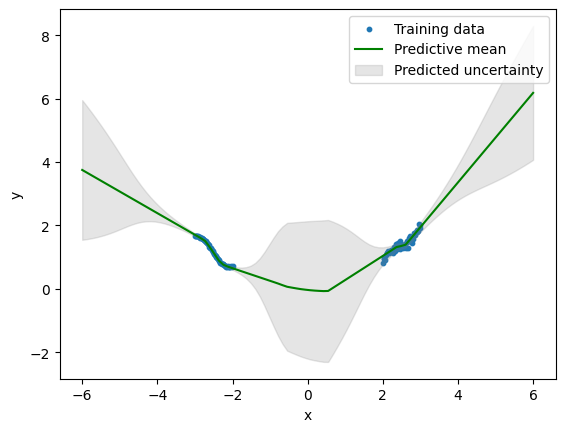}
        \label{fig:arbitraryVariance_high}
        \caption{High variance outside the training data}
    \end{subfigure}
    \hfill
    \caption{1-D constrained regression demonstrations with variance constraints outside the training data}
    \label{fig:arbitraryVariance}
\end{figure}

In this section, the effects of incorporating additional constraints, specifically value and derivative constraints, into the 1D regression model are demonstrated. These constraints are particularly focused on areas outside the training data range. Several experiments to understand how these constraints influence the model's predictions are conducted, especially in regions where training data is sparse or absent.
\paragraph{Value Constraints Outside Training Range:}
In the first demonstration, the two value constraints are placed at \(x = 5.0\) and \(x = 7.5\), both outside the training data range. The loss function for this is obtained as the sum of \(L_{VI}\) and the infeasibility constraints for both points. The infeasibility is calculated using:
\begin{equation}
\mathcal{L}_{\text{val}} =  E[y(x)] - y_{\text{gt}}(x)
\end{equation}
This setup allows us to observe the model's response to multiple external constraints and their impact on the overall prediction accuracy. The results, shown in Figure \ref{fig:1d_regressionPlots_a}, show that the proposed approach is able to accurately capture the value constraint. The compliance of the network to the constraints is strongly satisfied compared to the data, owing to the representational capacity of the model. 
\paragraph{Derivative constraints}
The derivative constraint is implemented by using central difference \(\delta\) around a small neighborhood \(\epsilon\) of the point (or region) in consideration:
\begin{equation}
\mathcal{L}_{\text{derivative}} = E[\delta_{\epsilon}(y(x))] - y_{\text{gt}}(x)
\end{equation}
It is seen in Figure~\ref{fig:1d_regressionPlots_d} that this approach is sufficient to capture derivative information. There is a caveat here that the 
\paragraph{Variance constraints}
The BE technique can encode any type of information as a constraint. This includes constraints on the predictive variance. While this is generally modulated by the data using the Gaussian likelihood term, it is possible to enforce constraints on its behavior given expert knowledge about it. Examples of such cases might be to specify the level of decay in prediction confidence outside the training data range. 
The variance constraint loss term is:
\begin{equation}
\mathcal{L}_{\text{var}} = \lambda (\sigma^2(x) - \sigma^2_{\text{gt}}(x))
\end{equation}
Figure~\ref{fig:arbitraryVariance} showcases three such examples of variance constraints on the output variance neuron, applied in regions outside the training data, with increasing uncertainty.
\paragraph{Conflicting Constraints}
Model behavior in the presence of conflicting constraints is shown in Figure \ref{fig:1d_regressionPlots_b}. Two different value constraints are applied at the same location \(x = 7.5\). This experiment is designed to evaluate how the model handles conflicting information and to understand the role of variance constraints in such situations. The variance at the constraint location is meant to express the uncertainty in the data provided in the form of a constraint. It is also reasonable to want to express arbitrary values of uncertainty at such points. However, the model needs to account for the variance that is admissible in these situations, so the predictive variance is constrained to be proportionate to the uncertain value constraints. 
\paragraph{Bound Constraints}
Next, an evaluation of the model with bound constraints to ensure that the predictions fall within a certain range is shown. For a given input \( x \), the bound constraints are defined by two functions, representing the upper bound \( \text{ub}(x) \) and lower bound \( \text{lb}(x) \). These are given by:
\begin{equation}
\text{ub}(x) = 1.0, \quad \text{lb}(x) = 0.5, \forall x \epsilon [-0.5, 0.5]
\end{equation}
where \( x \) is the input to the model.
\( n \) bound constraints are enforced, each represented as a hard constraint. For each test input \( x_i \in X_{\text{test}} \), the constraints are formulated as:
\begin{equation}
\text{lb}(x_i) \leq \text{pred}_i \leq \text{ub}(x_i), \quad \forall i \in \{1, \ldots, n\}.
\end{equation}
Like the equality constraint, the infeasibility, or violation, of the bounds is quantified by how far a prediction is from satisfying the constraints. The bound constraints are assessed by computing the deviation of the bounds from the original prediction:
\begin{equation}
L_{bound} = 
\begin{cases}
\text{min} - f(x) & \text{if } f(x) < \text{min} \\
0 & \text{if } \text{min} \leq f(x) \leq \text{max} \\
\text{max} - f(x) & \text{if } f(x) > \text{max}
\end{cases}
\end{equation}
The variance constraints modify the predicted aleatoric variance to minimize the absolute difference between the predicted and expected variance values.
\par
\begin{figure}[htbp]
    \centering
    % First row
    \begin{subfigure}[b]{0.49\textwidth}
        \includegraphics[width=\textwidth]{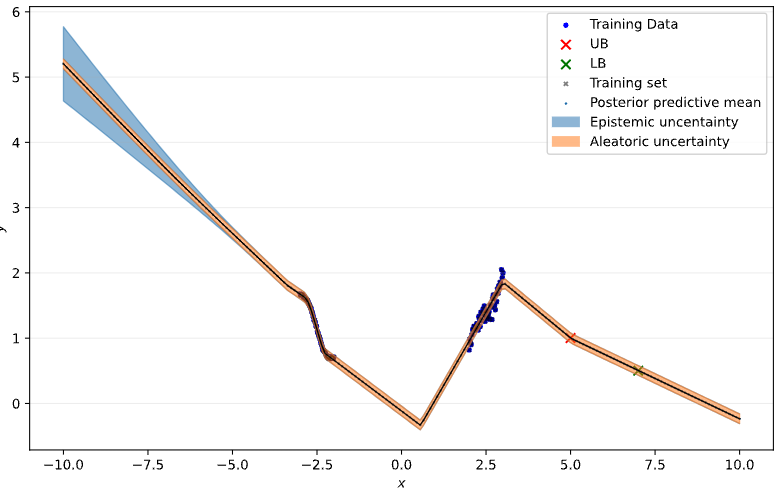}
        \caption{Value constraints at x = {5, 7.5}}
        \label{fig:1d_regressionPlots_a_PRBNN}
    \end{subfigure}
    \hfill
    \begin{subfigure}[b]{0.49\textwidth}
        \includegraphics[width=\textwidth]{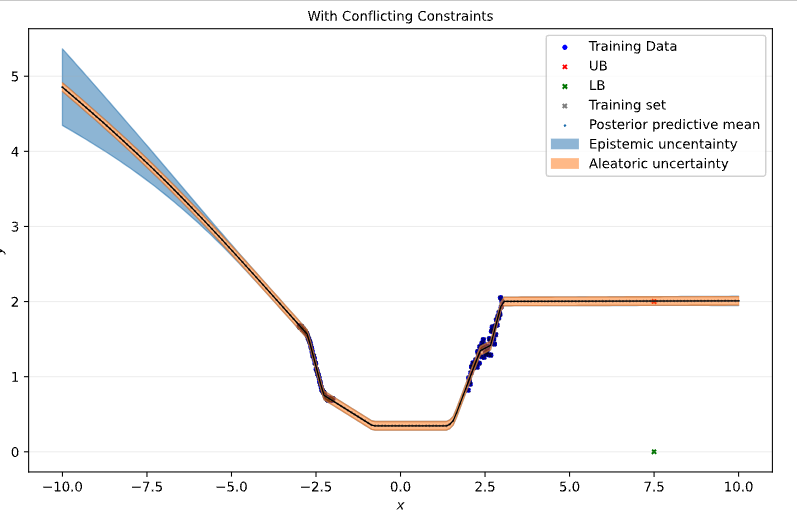}
        \caption{Conflicting value constraints at x = 7.5}
        \label{fig:1d_regressionPlots_b_PRBNN}
    \end{subfigure}
    \hfill
    % Second row
    \begin{subfigure}[b]{0.49\textwidth}
        \includegraphics[width=\textwidth]{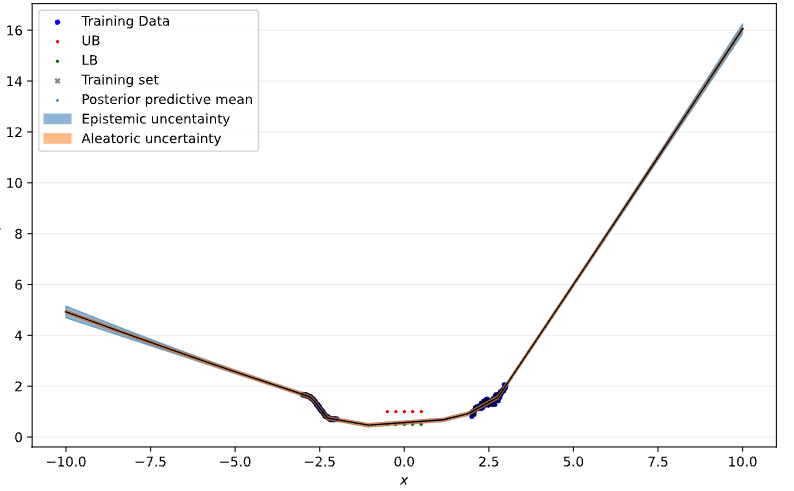}
        \caption{Bound constraint in x = [-0.5, 0.5]}
        \label{fig:1d_regressionPlots_c_PRBNN}
    \end{subfigure}
    \hfill
    \begin{subfigure}[b]{0.49\textwidth}
        \includegraphics[width=\textwidth]{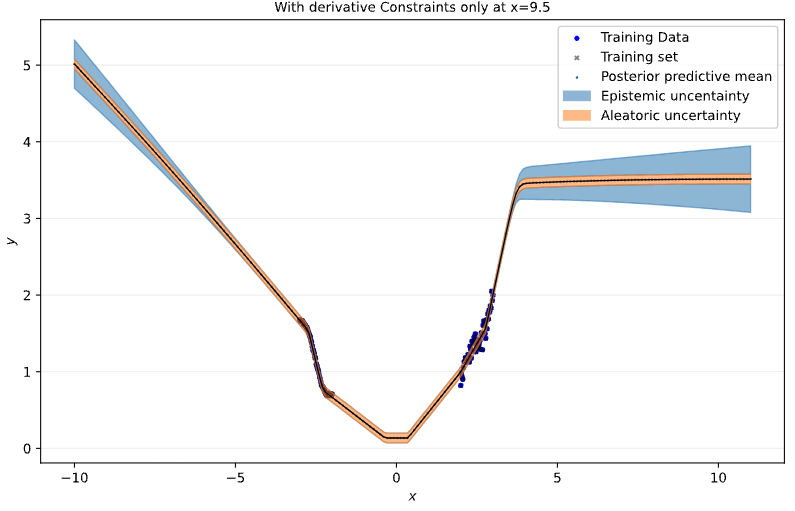}
        \caption{Derivative constraint \(\frac{dy}{dx} = 0, x = {9.5}\)}
        \label{fig:1d_regressionPlots_d_PRBNN}
    \end{subfigure}
    \hfill
    \caption{1-D constrained regression demonstrations using PR-BNN with value, derivative and bound constraints}
    \label{fig:1d_regressionPlots_PRBNN}
\end{figure}

Table~\ref{tab:comparison_regression} showcases a comparison of evaluation scores produced by PR-BNN \citep{Jiayu_2022} and BENN on constrained regression experiments. The evaluation scores are computed as the absolute value of the constraint violations, \(|y_{pred} - y_{gt}|\), where \(y_{pred}\) is averaged from 250 evaluations of the model. In the table, Inf-1, Inf-2 indicate the constraint violations of each constraint. For instance, the value constraints for both the conflicting and non-conflicting cases incorporate two separate constraints that need to be satisfied. In the case of the derivative constraint, we only have one constraint, whose violation is listed under Inf-1. Finally, for the bound constraint, the violation is averaged through the constraint domain points and listed under Inf-1. Figure~\ref{fig:1d_regressionPlots} and Figure~\ref{fig:1d_regressionPlots_PRBNN} shows a visual comparison of these two approaches. BENN shows significant differences with PR-BNN in the way conflicting constraints are handled. As a result, PR-BNN biases its prediction towards one of the two value constraints, whereas the BENN prediction occurs in between the two specified value constraints. The addition of the variance constraints in this work also allows experts to express uncertainties in the region of the specified constraints. The results of the derivative and bound constraints, however, do not show significant differences. It is to be noted, however, that the PR-BNN method does have lower constraint violations than the BENN method. 

\begin{table}[h]
\centering
\caption{Comparison of BENN and PR-BNN methods for value, bound and derivative constraints}
\label{tab:comparison_regression}
\begin{tabular}{|c|c|c|c|c|c|c|}
\hline
\multicolumn{1}{|c|}{Experiment} & \multicolumn{2}{c|}{BENN} & \multicolumn{2}{c|}{PR-BNN} \\ \hline
& Inf-1 & Inf-2 & Inf-1 & Inf-2 \\ \hline
Value Constraints - No Conflict & 0.023 & 0.041 & 0.001 &   0.038\\
Conflicting Value Constraint & 1.017 & 0.982 & 0.001& 2.001 \\
Derivative Constraint & 0.0016 & - & 4.216e-06& -\\
Bound Constraint & 0.0 & - & 0.0&- \\
\hline
\end{tabular}
\end{table}

\subsection{Beam Deflection Problem}
In this section, a classical problem in structural engineering is addressed: predicting the deflection of a beam under a given load. The data for this experiment is obtained from the theoretical deflection of a beam subjected to a load. The deflection of a beam, \( y \), as a function of its position, \( x \), is given by the function \(y(x)\), where the deflection depends on the material's Young's modulus (\( E \)), the length of the beam (\( L \)), the moment of inertia of the beam's cross-section (\( I \)), and the applied load (\( P \)). Specifically, the deflection function is defined piecewise for a beam of length \( 2L \) as:
\begin{equation}
y(x) = 
\begin{cases}
\frac{-P}{8EI}x^3 + \frac{PL}{4EI}x^2, & \text{if } 0 \leq x \leq 2L \\
\frac{P}{6EI}x^3 - \frac{3PL}{2EI}x^2 + \frac{7PL^2}{2EI}x - \frac{7PL^3}{3EI}, & \text{otherwise}
\end{cases}
\end{equation}
In this experiment, the values for \( E \), \( L \), \( I \), are set to typical values found in steel beams and \( P \) is set to 2000 N. For training purposes, a small sample of 10 points is observed by sampling points between \( 0.5L \) and \( 1.3L \). To simulate real-world measurements, a small Gaussian noise \(N(0,0.001)\) is added to the deflection values, resulting in the observed deflections, \( y_{\text{train}}(x) \). It should be noted that the problem setting indicates that the system is only partially observable (i.e., between \( 0.5L \) and \( 1.3L \)), and the other portion of the beam is non-observable. This is not uncommon in engineering practice due to the inaccessibility of the structure for equipment/sensors or due to the limited number of sensors/resources to measure the entire structure. 

The Bayesian Neural Network (BNN) used for regression predicts both the mean and the variance of the target variable. This network has an input layer with 1 neuron (representing the position \( x \)), and an output layer with 2 neurons, outputting both the predicted mean deflection and the log-transformed noise level. Further, it has 1 hidden layer, consisting of 2048 units. The GELU activation function \(x*\Phi(x)\) is used for this task, where \(\Phi(x)\) is the Cumulative Distribution Function (CDF) of the Gaussian. This activation helps significantly improve performance in the mean sense for the test set, evaluated from \(L = 0\) to \(L = 3\). 
\paragraph{Constraint Implementation}
To enhance the performance of the BENN model, specific constraints are applied where the direct measurements are not available due to the reasons mentioned above. These constraints are formulated to ensure that the model's predictions adhere to known physical principles and boundary conditions associated with beam deflection. Two types of constraints are implemented: value constraints and derivative constraints.
\begin{enumerate}
    \item \textbf{Value Constraint:} The value constraint is applied to ensure that the predicted deflection at specific points matches known values. This is particularly relevant at the beam's left end and the pin support at \(x = 2.0\), where the deflection is zero:
    \begin{equation}
        f(x = 0.0;\theta) = 0, \quad f(x = 2.0;\theta) = 0.
    \end{equation}
    \item \textbf{Derivative Constraint:} The derivative constraint is applied to ensure that the rate of change of the deflection at a certain point is known. In this case, the derivative is enforced to be zero at the point \( x = 0.0 \), corresponding to the beam's left end. Using finite differences at \( x = 0.0 \), we constrains this derivative to be zero. 

\end{enumerate}
\begin{figure}[htbp]
    \centering
    \begin{minipage}[b]{0.49\textwidth}
        \includegraphics[width=\textwidth]{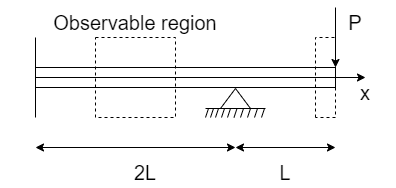}
        \caption{Beam configuration. The observable regions are given in the dashed boxes.}
        \label{fig:beamConfig}
    \end{minipage}
    \hfill % adds horizontal space between the two figures
    \begin{minipage}[b]{0.49\textwidth}
        \includegraphics[width=\textwidth]{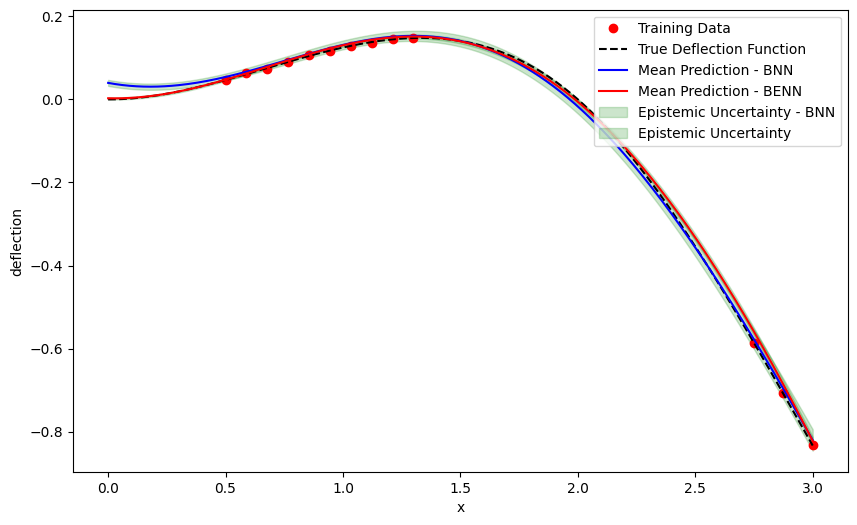}
        \caption{Predicted beam deflection with BENN and BNN.}
        \label{fig:beamDeflectionResults}
    \end{minipage}
\end{figure}
Figure~\ref{fig:beamDeflectionResults} shows that the proposed BENN model prediction significantly outperforms the BNN model. 
\subsection{Microstructure Generation}
\label{subsec: Microstructure_VAE}
% Overview of microstructures
In material science, predicting the properties of heterogeneous materials such as alloys, composites, polymers, and porous media is critical. These materials exhibit complex microstructures that influence their mechanical, thermal, electromagnetic, and other physical properties. Thus, efficient and scalable microstructure generation methods are critical for the structure-property analysis in material science. For example, \citet{gao2021ultra} proposes a method using a mixture random field model to generate a non-Gaussian field. This statistical method can generate binary phase microstructures that can either be anisotropic or isotropic and require the specification of two-point correlation functions. This approach belongs to the statistical descriptor approach and has an interpretable physical meaning of the generated microstructure. However, this approach can only handle second-order statistical descriptors and cannot generate some microstructures that show very complex patterns and high-order statistics. 
Alternatively, Neural Network-based microstructure generation can handle arbitrary complex microstructure patterns due to the universal approximation nature of NNs. For example, autoencoders, leveraging advancements in machine learning, provide a modern computational tool that offers significant benefits in the simulation and analysis of microstructures.

\par
Autoencoders facilitate the fast and efficient simulation of microstructures, reducing reliance on extensive experimental datasets and accelerating the material design process. This capability is crucial for designing materials with optimized properties tailored for specific applications, essential in sectors like aerospace, automotive, and biomedical engineering. Furthermore, the computational generation of microstructures allows for the quantification and management of the inherent uncertainties in materials, which is vital for developing reliable materials and implementing robust design principles.Moreover, generating microstructures using autoencoders fits seamlessly into broader computational frameworks such as Integrated Computational Materials Engineering (ICME), as discussed by \citet{gao2021ultra}. However, traditional autoencoders do not necessarily reproduce the known statistical descriptors from classical microstructure quantification. Thus, the motivation of the proposed autoencoder-based approach is aimed at incorporating the classical statistical descriptors as constraints in the NN architecture using the proposed BENN methodology.

% Discussion about VAE architecture
\subsubsection{Bayesian Entropy Convolutional Variational Auto-Encoder (BE-CVAE)}
The proposed architecture integrates a Convolutional Variational Autoencoder (CVAE) focusing on image generation while accommodating specific constraints using the BE framework. The encoder is a sequence of convolutional layers, each followed by a Rectified Linear Unit (ReLU) activation function, MaxPooling for spatial downsampling, and Batch Normalization. The convolutional layers extract hierarchical features to finally obtain the latent space \(Z\).
\par
The latent space is characterized by two fully connected layers, one for the mean $\mu$ and the other for the log-variance $\log(\sigma^2)$, of a Gaussian distribution. The model uses the reparameterization trick, generating a sample $z$ from the latent space by 
where $\epsilon$ is a random noise sampled from a standard normal distribution.
\par
The decoder reconstructs the input image using transposed convolutional layers with ReLU activations and Batch Normalization. A Sigmoid function in the final layer outputs pixel values in the [0,1] range. Post-decoding, the output is resized to the original input dimensions using interpolation.
\subsubsection{Dataset}
The dataset is generated using a non-Gaussian random field, without using techniques such as simulated annealing, which are computationally intensive.  The use of this technique allows for the simultaneous generation of both the microstructure and the latent material property field. Using this approach, up to 150 training samples are generated for the microstructure generative model experiments. 
% Discussion about constraints
\subsubsection{Constraint - Two-point correlation function}
The Two Point Correlation Function (TPCF) serves as a fundamental constraint in the analysis and synthesis of binary microstructure images. This statistical tool quantifies the spatial correlation between pairs of points within a given distance in a microstructure, providing insight into its heterogeneity and spatial distribution features.
In the proposed model, the TPCF is calculated for each generated microstructure image. The function evaluates the probability of finding two points, separated by a certain distance, that are both in the same state (either material or void). The TPCF constraint ensures that the generated microstructures possess spatial characteristics similar to those of the training set. This includes features like the distribution of phases, the degree of homogeneity or clustering, and the overall geometric properties of the material. By applying this constraint, it is aimed to produce microstructures that are not only statistically representative of the real material system but also maintain critical physical and mechanical properties. 

Ensuring fidelity to the TPCF is particularly significant in materials science and engineering, where the microstructural arrangement greatly influences the macroscopic properties of materials. The accurate reproduction of TPCF in synthetic microstructures is thus essential for reliable material property predictions and subsequent applications.
\par
The Two Point Correlation Function (TPCF) in this study is computed using a Fourier Transform Autocorrelation approach, which is an efficient method for analyzing spatial patterns in microstructures. This method is based on the principle that the Fourier transform of an autocorrelation function is the squared magnitude of the Fourier transform of the original function. The steps of the computation, described in \citep{Suankulova2020} are summarized as follows:
\begin{enumerate}    
    \item \textbf{Fourier Transform:} The Fourier Transform of the binary microstructure is computed. This step is facilitated by the use of the Fast Fourier Transform (FFT), which is computationally efficient for digital images.
    
    \item \textbf{Power Spectrum:} Next, the power spectrum of the Fourier-transformed image is obtained by calculating its squared magnitude. This power spectrum reflects the frequency components of the spatial pattern of the microstructure.
    
    \item \textbf{Inverse Fourier Transform:} The inverse Fourier Transform is applied to the power spectrum. This operation converts the frequency-domain representation back into the spatial domain, resulting in an autocorrelation function.
    
    \item \textbf{Radial Averaging:} Finally, for the TPCF, radial averaging is performed on the autocorrelation function. This process involves averaging the values over circles (in 2D) of radius \( r \), which provides the final TPCF.
\end{enumerate}
This Fourier Transform approach to computing the TPCF is advantageous due to its computational efficiency, particularly for large and complex microstructures. It also inherently accommodates periodic boundary conditions, which is beneficial in materials with repeating structures.

The resultant TPCF from this method provides insights into the heterogeneity and spatial distribution of features within the microstructure, which is crucial for understanding how these features influence the material's macroscopic properties.
\par
To facilitate gradient-based optimization in the process of image binarization, a differentiable binarization function is employed, defined as follows:
\begin{equation}
    \text{b}(x) = \sigma(s \cdot (x - \theta)),
\end{equation}
where \( \sigma \) denotes the sigmoid function, \( s \) is the steepness parameter, \( \theta \) is the threshold, and \( x \) is the input value. 
The sigmoid function \( \sigma(x) = \frac{1}{1 + e^{-x}} \) is used to provide a smooth transition between 0 and 1. The steepness parameter \( s \), set to a high value (e.g., 100), ensures a sharp transition around the threshold \( \theta \). This approach allows for an approximate binary representation while retaining differentiability, which is essential for gradient-based optimization methods.

The TPCF is now ready to be used as a functional constraint. This amounts to either solving multiple value constraints for the TPCF sequentially, or to compute the overall absolute deviation from the expected curve at once. The latter option is computationally cheaper than evaluating multiple value constraints per iteration.
\subsubsection{Constraint - Porosity}
The porosity of a microstructure is a critical characteristic because it directly influences the material's mechanical, thermal, and transport properties. High porosity can lead to reduced strength and increased permeability, which are important factors in applications such as structural components and filtration systems. Accurate control and prediction of porosity are important for material design workflows that have specific performance criteria. The porosity function is given by:
\begin{equation}
    e = \frac{V_p}{V_p + V_s},
\end{equation}
where \( V_p \) is the total number of voxels in the void phase and \( V_s \) is the total number of voxels in the solid phase.
\subsubsection{Results}
The proposed model was first tested to compare how fast convergence occurred in generating quality samples. The baseline case with no constraints took far longer to converge compared to the ones using constraints. Furthermore, the model that used both porosity and the TPCF constraint converged significantly faster than both the baseline and the model that only incorporated porosity constraints. Next, the effect of training data on performance is studied. The performance of the generative model is measured by how well the generated structures conform to the expected TPCF function. Table~\ref{tab:BENN_Microstructure_TrainingData} shows strong evidence of improvement in compliance with the TPCF requirements as more constraints are added. As expected, an improvement is also seen when the number of training samples are increased. It is to be noted, however, that the evolution of constraint compliance during training paints a more nuanced picture in high dimensional problems. The resulting absolute errors from the TPCF and porosity constraints are lower those seen in the baseline model, it is still non-zero. This is because the TPCF and porosity constraints end up guiding the model to produce faster convergence with better reconstruction adherence with these two constraints. 

\begin{figure}
    \centering
    \includegraphics[width=\textwidth]{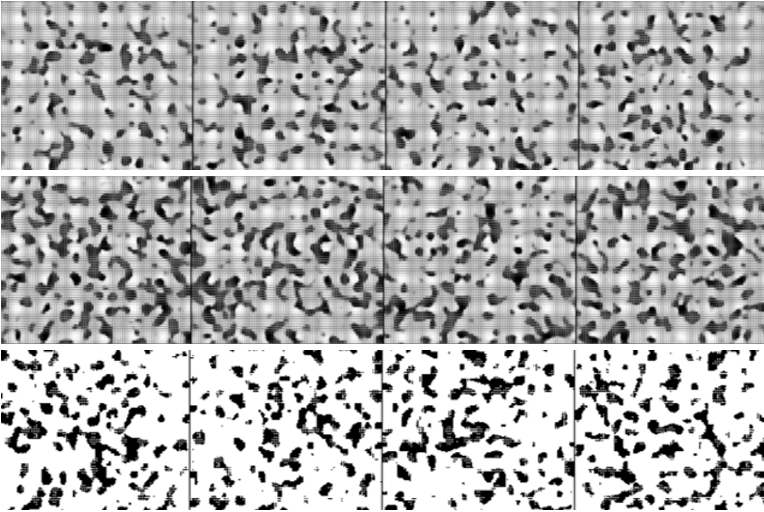}
    \caption{Microstructure generation using the proposed model after training the model with 260 epochs with (Top Row) No constraints (the baseline CVAE model) (Middle Row) TPCF Constraint (Bottom Row) TPCF + Porosity Constraints. Note that the model converges faster with two constraints compared to the baseline or just using the TPCF constraint}
    \label{fig:BENN_MicrostructureDemo}
\end{figure}

% first result needs to compare the CVAE with the BE-CVAE across training samples
\begin{table}[h!]
\centering
\begin{tabular}{|c|c|c|c|}
\hline
\textbf{Training samples} & \textbf{No constraint} & \textbf{TPCF} & \textbf{TPCF + Porosity} \\ \hline
25 & 0.85 & 0.67 & 0.60 \\ \hline
50 & 0.59 & 0.46 & 0.38 \\ \hline
100 & 0.59 & 0.41 & 0.39 \\ \hline
150 & 0.47 & 0.29 & 0.34 \\
\hline
\end{tabular}
\caption{TPCF Constraint compliance as L1 error on generated samples: Analysis along training data and number of constraints}
\label{tab:BENN_Microstructure_TrainingData}
\end{table}
%%%%%%%%%%%%%%%%%%%%%%%%
% \newpage
% \section{Discussion}
% \label{sec:Discussion}
\newpage
\section{Conclusion}
\label{sec:Conclusion_BENN}
In this paper, the Bayesian Entropy Neural Networks (BENN) framework was introduced, and its application in various settings, including 1D regression, beam deflection modeling, and microstructure generation, was demonstrated. The integration of the Maximum Entropy (MaxEnt) principle with the Bayesian framework allowed for the imposition of constraints on predictions, thereby offering a novel approach to constrained deep learning. Note that this method is a general framework that explains the approaches used in similar works such as \citep{Jiayu_2022} and \citep{Yang_2020_Interpretable}.
\par
The experiments with 1D regression emphasized the BENN framework’s versatility in handling value, derivative and variance constraints. It was observed that the model could effectively accommodate external constraints, even in areas with sparse or absent training data. This was particularly evident in experiments involving conflicting constraints and variance control, illustrating BENN’s ability to manage complex and uncertain information. The model can, therefore, be used for problems that require adherence to boundary conditions through constraints. An interesting demonstration of BENN in this work was in the domain of microstructure generation. The integration of constraints like the Two-Point Correlation Function (TPCF) and porosity into a Convolutional Variational Autoencoder showed substantial improvements in generating microstructures that are not only statistically representative but also improve adherence to critical physical properties. This highlights the framework's potential in materials science, where accurate microstructure characterization is crucial.
\par
This paper provides some areas for improvement and further investigation. The computational complexity associated with performing simultaneous optimization of model parameters and Lagrangian multipliers is a challenge, particularly for large-scale problems. Another avenue for advancement lies in the integration of epistemic uncertainty constraints within the BENN framework. Epistemic, or model, uncertainty arises from a lack of knowledge about the best model to represent a process. In many real-world applications, especially in fields like climate modeling and epidemic modeling, understanding and quantifying epistemic uncertainty is crucial for making reliable predictions and decisions. Incorporating these constraints could enhance the model's ability to express uncertainty in predictions, especially in scenarios with limited data or where the data does not capture the entire spectrum of the underlying distribution. This could involve developing new methodologies for characterizing and quantifying epistemic uncertainty and devising novel ways to embed this information into the learning process. Finally, papers that work on applying the BENN framework to a wider array of practical applications would not only demonstrate its versatility but also uncover specific challenges and opportunities for improvement.

\section{Acknowledgment}
The work was partially supported by funds from the National Science Foundation (Award Number: 2331781) and by funds from Arizona State University. The support is greatly acknowledged. 

\bibliographystyle{elsarticle-harv}  
\bibliography{cas-refs}

\end{document}